\documentclass[runningheads]{llncs}
\usepackage[T1]{fontenc}
\usepackage{graphicx}
\usepackage{cite}
\usepackage{amsmath,amssymb,amsfonts}
\usepackage{algorithmic}
\usepackage{graphicx}
\usepackage{textcomp}
\usepackage{balance}
\usepackage{multirow}
\usepackage[table,xcdraw]{xcolor}
\usepackage{comment}
\usepackage{tikz}
\usepackage{amssymb}
\usepackage{pifont}
\usepackage[misc,geometry]{ifsym}
\usepackage{nicefrac}

\begin{document}

\title{On the Uncertain Single-View Depths in Colonoscopies}
%
%

\author{Javier Rodriguez-Puigvert{$^{\textrm{(\Letter)}}$} \and David Recasens \and Javier Civera \and Ruben Martinez-Cantin}
\authorrunning{J.Rodriguez-Puigvert et al.}
%
\institute{Universidad de Zaragoza, Zaragoza, Spain \email{\{jrp,recasens,jcivera,rmcantin\}@unizar.es}}
\maketitle              

\begin{abstract}
Estimating depth information from endoscopic images is a prerequisite for a wide set of AI-assisted technologies, such as accurate localization and measurement of tumors, or identification of non-inspected areas. As the domain specificity of colonoscopies --deformable low-texture environments with fluids, poor lighting conditions and abrupt sensor motions-- pose challenges to multi-view 3D reconstructions, single-view depth learning stands out as a promising line of research. Depth learning can be extended in a Bayesian setting, which enables continual learning, improves decision making and can be used to compute confidence intervals or quantify uncertainty for in-body measurements. In this paper, we explore for the first time Bayesian deep networks for single-view depth estimation in colonoscopies.
Our specific contribution is two-fold: 1) an exhaustive analysis of scalable Bayesian networks for depth learning in different datasets, highlighting challenges and conclusions regarding synthetic--to--real domain changes and supervised vs. self-supervised methods; and 2) a novel teacher-student approach to deep depth learning that takes into account the teacher uncertainty.

\end{abstract}

\section{Introduction}
\label{sec:Introduction}
Depth perception inside the human body is one of the cornerstones to enable automated assistance tools in medical procedures (e.g. virtual augmentations and annotations, accurate measurements or 3D registration of tools and interest regions) and, in the long run, the full automation of certain procedures and medical robotics. Monocular cameras stand out as very convenient sensors, as they are minimally invasive for in-vivo patients, but estimating depth from colonoscopy images is a challenge.
Multi-view approaches are accurate and robust in many applications outside the body, e.g.~\cite{schonberger2016structure}, but assume certain rigidity, texture and illumination conditions that are not fulfilled in in-body images. Single-view 3D geometry is ill-posed, since infinite 3D scenes can explain a single 2D view~\cite{Hartley:2003:MVG:861369}. In this last case, deep neural networks have shown impressive results in last years~\cite{eigen2014depth,fu2018deep,godard2019digging}. However, the vast majority of deep learning models lack any metric or intuition about their predictive accuracy. In a critical environment such as the inside of the human body, uncertainty quantification is essential. Specifically, in medical robotics it allows us to properly account for uncertainty in control and decision making, and in SLAM~\cite{1638022} to safely navigate inside the body. It also provides confidence intervals in in-body measurements (e.g., polyps), which is valuable for doctors to decide how to act. Uncertainty quantification is, in general, a must-have for robust, interpretable and safe AI systems. Bayesian deep learning perfectly combines the fields of deep learning and uncertainty quantification in a sound and grounded approach. However, for high-dimensional deep networks, accurate Bayesian inference is intractable. Only bootstrapping methods such as deep ensembles~\cite{Lakshminarayanan2017} have shown to produce well-calibrated uncertainties in many computer vision tasks at reasonable cost~\cite{Gustafsson2019}. 

\begin{figure}[t!]
\centerline{\includegraphics[width=\columnwidth,keepaspectratio]{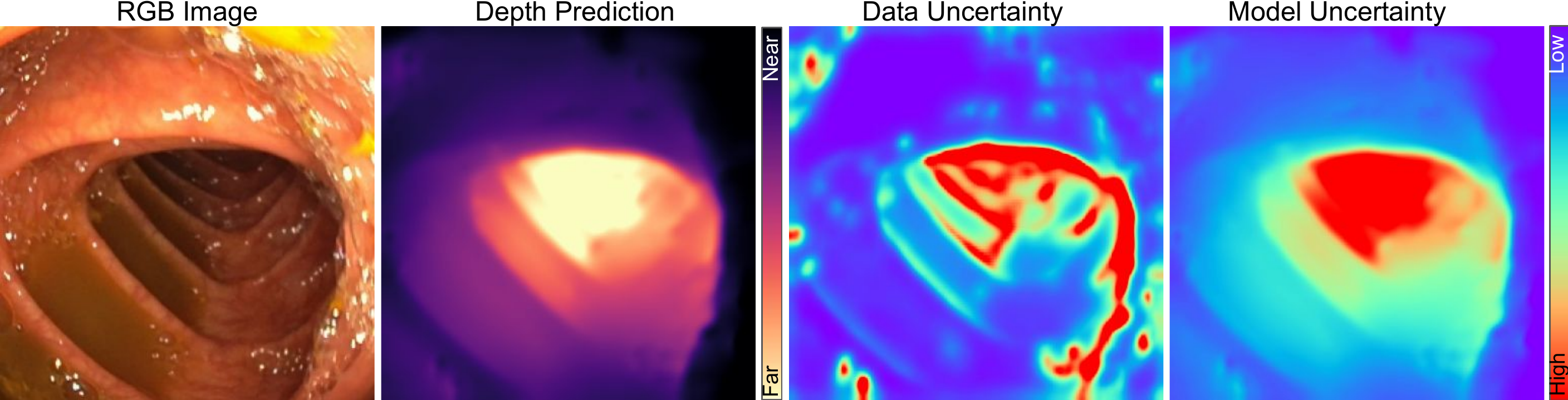}}
\caption{Depth and uncertainty predictions for a colonoscopy image. Dark/bright colors stands for near/far depths, blue/red stands for low/high uncertainties. Note the higher uncertainties in darker and farther areas and in reflections.}
\label{fig:teaser}
\end{figure}

In this work, we address for the first time the use of Bayesian deep networks for depth prediction in colonoscopies. Fig.~\ref{fig:teaser} illustrates the predicted depth and uncertainties of one of our Bayesian model in a real colonoscopy image. Specifically, we first benchmark thoroughly, in synthetic and real data, supervised and self-supervised learning approaches in the colonoscopic domain.
We demonstrate that Bayesian models trained on synthetic data can be transferred adequately to similar domains and we quantify the generalization of depth and uncertainty to real scenarios. 
Secondly, we propose a novel teacher-student method that models the uncertainty of the teacher in the loss, achieving state-of-the-art performance in a real colonoscopy domain.

\section{Preliminaries and Related Work}
\label{sec:Related_work}
\textbf{Bayesian deep learning}
is a form of deep learning that performs probabilistic inference on deep network models. This enables uncertainty quantification for the model and the predictions. For high-dimensional deep networks, Bayesian inference is intractable and some approximate inference methods such relying on variational inference or Laplace approximations might perform poorly. In practice, sampling methods based on bootstrapping, such as deep ensembles \cite{Lakshminarayanan2017}, or Monte Carlo, like MC-dropout \cite{Gal2016}, have shown to be the most scalable, reliable and efficient approaches for depth estimation and other computer vision tasks \cite{Gustafsson2019}. 
In particular, deep ensembles have shown to perform extremely well even with a reduced number of samples, because each random sample of the network weight is optimized using a maximum a posteriori (MAP) loss $\mathcal{L}_{MAP}=\mathcal{L}_{LL} + \mathcal{L}_{prior}$, resulting in a high probability sample.  The MAP loss requires a prior distribution, which unless otherwise stated, we assume to be a Gaussian distribution over the weights $\mathcal{L}_{prior} = ||\theta||^2$. For the data likelihood $\mathcal{L}_{LL}$, we use a loss function based on the Laplace distribution for which the predicted mean $\mu(x)$ and the predicted scale $\sigma(x)$ come from the network described in Section \ref{sec:supervisedlearning}, with two output channels \cite{Kendall2017}. The variance associated with the scale term represents the uncertainty associated with the data, also called \emph{aleatoric uncertainty} or $\sigma_a(x)$. Furthermore, in deep ensembles, the variance in the prediction from the multiple models of the ensemble is the uncertainty that is due to the lack of knowledge in the model, which is also called \emph{epistemic uncertainty} or $\sigma_e(x)$. For example, data uncertainty might appear in poorly illuminated areas or with lack of texture, while model uncertainty arises from data that is different from the training dataset. Note that while the model uncertainty can be reduced with larger training datasets, data uncertainty is irreducible. 
Model uncertainty is particulary relevant to address domain changes. To illustrate that, in Section \ref{sec:Experimental_results} we present results of models trained on synthetic data and tested on real data. \\ \\ \noindent \textbf{Single-View Depth Learning} has demonstrated a remarkable performance recently. Some methods rely on accurate ground truth labels at training \cite{eigen2014depth,fu2018deep,song2021monocular}, which is not trivial in many application domains. Self-supervision without depth labels was achieved by enforcing multi-view photometric consistency during training \cite{zhou2017unsupervised, zhan2018unsupervised,godard2019digging}.
In the medical domain, supervised depth learning was addressed by Visentini et al.~\cite{visentini2017deep} with autoencoders and by Shen et al.~\cite{shen2019context} with GANs, both using ground truth from phantom models.
Other works based on GANs were trained with synthetic models~\cite{chen2019slam,mahmood2018unsupervised,mahmood2018deep,rau2019implicit}, and Cheng et al.~\cite{10.1007/978-3-030-87231-1_12} added a temporal consistency loss.
Self-supervised learning is a natural choice for endoscopies to overcome the lack of depth labels on the target domain \cite{sharan2020domain,ozyoruk2021endoslam,recasens2021endo}. 
Although depth or stereo are not common for in-vivo procedures, several works use them for training \cite{luo2019details,xu2019unsupervised,huang2021self}. Others train in phantoms \cite{turan2018unsupervised} or synthetic data \cite{freedman2020detecting,hwang2021unsupervised}, facing the risk of not generalizing to the target domain. In this paper we study the limits of such generalization. SfM supervision was addressed by Liu et al.~\cite{liu2019dense} using siamese networks and by Widya et al.~\cite{widya2021self} using GANs. Note that none of these references address uncertainty quantification, which we cover in this work.Kendall et al. \cite{Kendall2017} combine epistemic and aleatoric uncertainty by using a MC Dropout approximation of the posterior distribution. This approach obtains pixel-wise depth and uncertainty predictions in a supervised setting.
Ilg et al. \cite{ilg2018uncertainty} propose a multi-hypothesis network to quantify the uncertainty of optical flow.
Traditional self-supervised losses to regress depth are limited due to the aleatoric uncertainty of input images \cite{li2021sins}. 
Poggi et al.~\cite{poggi2020uncertainty} address such problem by introducing a teacher-student architecture to learn depth and uncertainty. As key advantages, teacher-student architectures provide aleatoric uncertainty for depth and avoid photometric losses and pose regression networks, which are frequently unstable. 
 In this work, we evaluate supervised and self-supervised approaches in colonoscopies images and propose a novel teacher-student approach that includes teacher uncertainty during training. Among the scalable Bayesian methods for single-view depth prediction, deep ensembles show the best calibrated uncertainty \cite{rodriguez2021bayesian, poggi2020uncertainty} and hence we choose them as our model.
\begin{figure}[t]
  \centering
\includegraphics[width=\columnwidth]{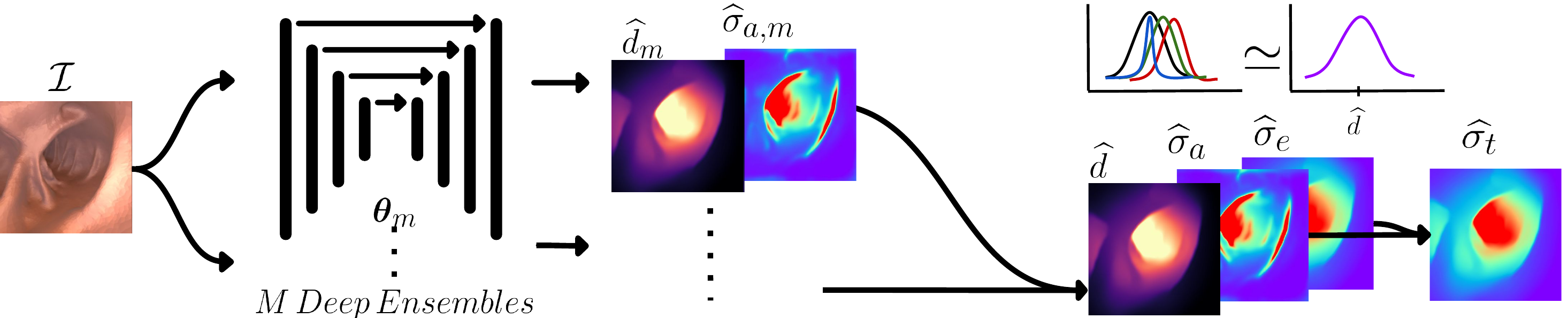}
  \caption{Forward propagation of supervised deep ensembles. Our deep ensembles model a Gaussian distribution $N(\widehat{d},\widehat{\sigma}^2_{t})$.  $\widehat{d}$ comes from averaging all ensembles depth output and $\widehat{\sigma}^2_{t}$ from joining data and model uncertainties}
\label{fig:testsup}
\end{figure}

\section{Supervised Learning using Deep Ensembles}
\label{sec:supervisedlearning}

Let our dataset $\mathcal{D}=\{ \{\mathcal{I}_1,d_1 \},\hdots,\{\mathcal{I}_N,d_N \}\}$ be composed by $N$ samples, where each sample $i \in \{1,\hdots,N\}$ contains the input image $\mathcal{I}_i \in \{0,\hdots,255\}^{w \times h \times 3}$ and per-pixel depth labels $d_i \in \mathbb{R}^{w \times h}_{>0}$. Regarding our network, we use an encoder-decoder architecture with skip connections, inspired by Monodepth2~\cite{godard2019digging}, with two output layers. Thus, for every new image $\mathcal{I}$ the network predits its pixel-wise depth $\widehat{d}(\mathcal{I}, \boldsymbol{\theta}) \in \mathbb{R}^{w \times h}_{>0}$ and data variance $\widehat{\sigma}_{a}^2(\mathcal{I}, \boldsymbol{\theta}) \in \mathbb{R}^{w \times h}_{>0}$. As commented in Section \ref{sec:Related_work}, we use a MAP loss, with $\mathcal{L}_{prior} = ||\boldsymbol{\theta}||^2$ and :

\begin{equation}
    \mathcal{L}_{LL} = \frac{1}{w \cdot h} \sum_{\boldsymbol{j} \in \Omega_{i}} \left(\frac{||d\!\left[ \boldsymbol{j} \right] - \widehat{d}\!\left[ \boldsymbol{j} \right]||_1}{ \widehat{\sigma}_{a}\!\left[ \boldsymbol{j} \right]} +\log \widehat{\sigma}_{a}\!\left[ \boldsymbol{j} \right] \right)
    \label{eq:losssupervised}
\end{equation}
where $\left[\cdot \right]$ is the sampling operator and $\boldsymbol{j} \in \Omega$ refers to the pixel coordinates in the image domain $\Omega$. The per-pixel depth labels $d$ can be obtained from ground truth depth $d^{GT}$ or from SfM 3D reconstructions $d^{SfM}$~\cite{schonberger2016structure}.
A \emph{deep ensemble model} is composed by $M$ networks with weights $\{\theta_m\}_{m=1}^M$, each of them trained separately starting from different random seeds. We denote as  $(\widehat{d}_m,\widehat{\sigma}^2_{a,m})$ the output of the $m^{th}$ ensemble (see Fig. \ref{fig:testsup}). We obtain the mean depth of the ensemble $\widehat{d}$ and its epistemic uncertainty $\widehat{ \sigma}^2_{e}$ using the total mean and variance of the full model.
The total uncertainty $\widehat{\sigma}^2_{t}  = \widehat{\sigma}^2_{a} + \widehat{\sigma}^2_{e}$ combines the data $\widehat{\sigma}^2_{a}$ and model $\widehat{\sigma}^2_{e}$ uncertainties which results from the law of total variance.
\begin{equation}
\widehat{d} = \frac{1}{M} \sum_{m=0}^{M} \widehat{d}_m, \ \  \widehat{\sigma}^2_{a} = \frac{1}{M} \sum_{m=0}^{M} \widehat{\sigma}^2_{a,m}, \ \ 
\widehat{\sigma}^2_{e} = \frac{1}{M} \sum_{m=0}^{M} \left(\widehat{d} - \widehat{d}_m\right)^2
\label{eq:supervisedL1}
\end{equation}
\section{Self-Supervised Learning using Deep Ensembles} 
Self-supervised methods aim at learning \emph{without} depth labels, the training data being $\mathcal{D}=\{ \mathcal{I}_1 ,\hdots, \mathcal{I}_N\}$ and the supervision coming from multi-view consistency.  
For each instance $m$ of a deep ensemble, two deep networks are used \cite{godard2019digging}. The first one learning depth and a photometric uncertainty parameter $\widehat{u}$ and the second one learning to predict relative camera motion. We use a pseudo-likelihood for the loss function, that uses both networks for photometric consistency:
\begin{equation}
    \mathcal{L}_{LL,m} = \frac{1}{w \cdot h} \sum_{\boldsymbol{j} \in \Omega_{i}} \left( \frac{\mathcal{F}_p\!\left[{\boldsymbol{j}}\right] }{ \widehat{u}_{m}\!\left[{\boldsymbol{j}}\right]} +\log \widehat{u}_{m}\!\left[{\boldsymbol{j}}\right]\right)
    \label{eq:unsupervised_loss}
\end{equation}
where $\mathcal{F}_p$ is the photometric residual and $\widehat{u}_{m}$ an uncertainty prediction. The photometric residual $\mathcal{F}_p\!\left[{\boldsymbol{j}}\right]$ of pixel $\boldsymbol{j}$ in a target image $\mathcal{I}_i$ is the minimum --between the warped images $\mathcal{I}_{i^{\prime} \rightarrow i}$ from the previous and posterior images $\mathcal{I}_{i^\prime}$ to the target one $\mathcal{I}_i$-- of the sum of the photometric reprojection error and Structural Similarity Index Measure (SSIM)~\cite{wang2004image}:

\begin{equation}
    \mathcal{F}_p\!\left[{\boldsymbol{j}}\right] = \min ( (1-\alpha) \lVert \mathcal{I}_{i}\!\left[{\boldsymbol{j}}] \! - \mathcal{I}_{i^{\prime} \rightarrow {i}}\![{\boldsymbol{j}}\right] \rVert_1 +
    \frac{\alpha}{2} (1 -  \text{SSIM}\! (\mathcal{I}_{i},\mathcal{I}_{i^{\prime} \rightarrow {i}},\boldsymbol{j}) \!)
\end{equation}
being $\alpha \in \left[0,1\right]$ the relative weight of the addends; and $\mathcal{I}_{i}\!\left[{\boldsymbol{j}}\right]$ and $\mathcal{I}_{i^{\prime} \rightarrow {i}}\!\left[{\boldsymbol{j}}\right] = \mathcal{I}_{i^\prime}\!\left[{\boldsymbol{j^\prime}}\right]$ the color values of pixel $\boldsymbol{j}$ of the target image $\mathcal{I}_{i}$ and of the warped image $\mathcal{I}_{i^{\prime} \rightarrow i}$. To obtain this latter term, we warp every pixel $\boldsymbol{j}$ from the target image domain $\Omega_{i}$ to that of the source image $\Omega_{i^{\prime}}$ using:
\begin{equation}
   {\boldsymbol{j}^{\prime} = \pi \! \left( \mathbf{R}_{i^{\prime} {i}} \pi^{\!-\!1}\!(\boldsymbol{j},\widehat{d}_{i}\!\left[{\boldsymbol{j}}\right] ) +  \mathbf{t}_{i^{\prime} {i}}\right)}
\label{eq:backandforthprojection}
\end{equation}
$\mathbf{R}_{i^{\prime} {i}} \! \in \! \ensuremath{\mathrm{SO}(3)}$ and $\mathbf{t}_{i^{\prime} {i}} \! \in \! \mathbb{R}^3$ are the rotation and translation from $\Omega_{i}$ to $\Omega_{i^{\prime}}$, and $\pi$ and $\pi^{-1}$ the projection and back-projection functions (3D point to pixel and vice versa).
In this case, the prior loss also incorporates an edge-aware smoothness term $\mathcal{F}_s$, regularizing the predictions \cite{godard2019digging}. 
Thus, the prior term becomes $\mathcal{L}_{prior} = ||\theta||^2 + \lambda_u \mathcal{F}_s\!\left[{\boldsymbol{j}}\right]$, where $\lambda_u$ calibrates the effect of the smoothness in terms of the reprojection uncertainty. This prior term is then combined to obtain $\mathcal{L}_{MAP}$ as described in Section \ref{sec:Related_work}.
We obtain the ensemble prediction by model averaging as in the supervised case (Eq. \ref{eq:supervisedL1}).
In this case, the data uncertainty for the depth prediction $\widehat{\sigma}^2_{a,m}$ cannot be extracted from the photometric uncertainty parameter $\widehat{u}$. Due to this, only model uncertainty will be considered in the experiments ($\widehat{\sigma}^2_{t}  = \widehat{\sigma}^2_{e}$).

\section{Teacher-Student with Uncertain Teacher}
In the endoscopic domain, accurate depth training labels can only be obtained from RGB-D endoscopes (which are highly unusual) or synthetic data (that is affected by domain change).
We propose the use a of a Bayesian teacher trained on synthetic colonoscopies that produces depth and uncertainty labels.
The teacher's epistemic uncertainty allows us to overcome the domain gap automatically. 
Specifically, our novel teacher-student architecture models depth labels from the predictive posterior of the teacher $d \sim \mathcal{N}(\widehat{d}, \sigma^2_T)$ ($\sigma^2_T$ is the total teacher variance). Thus, the likelihood must incorporate both the teacher and student distributions, which is used in the training loss. As before, the loss is based on a Laplacian likelihood
\begin{equation}
    \mathcal{L}_{LL,m} = \frac{1}{w \cdot h} \sum_{\boldsymbol{j} \in \Omega_{i}} \left( \frac{||\widehat{d}_{T}\!\left[{\boldsymbol{j}}\right] - \widehat{d}\!\left[{\boldsymbol{j}}\right]||_1}{\widehat{\sigma}_{m}\!\left[{\boldsymbol{j}}\right]} +  \log \widehat{\sigma}_{m}\!\left[{\boldsymbol{j}}\right] \right)
    \label{eq:teacher-student}
\end{equation}
where the per-pixel variance is the sum of the teacher predictive variance and the aleatoric one predicted by the student $\widehat{\sigma}_{m}^2 = \widehat{\sigma}_{T}^2 + \widehat{\sigma}_{a,m}^2$. Our student is hence aware of the label reliability, which will be affected by the domain change.
\label{sec:Experimental_results}
\section{Experimental Results}
We present results in synthetic and real colonoscopies. Our first dataset is the one generated by Rau et al.~\cite{rau2019implicit}, containing RGB images rendered from a 3D model of the colon in 15 different texture and illumination conditions. The second one, the EndoMapper dataset \cite{endommaper} contains real monocular colonoscopies.
We evaluate each method using the following depth error metrics \cite{eigen2014depth}: absolute relative error $\nicefrac{1}{w \cdot h} \sum_{\boldsymbol{j} \in \Omega_{i}}\nicefrac{|d[j]-\widehat{d}[j]|}{\widehat{d}[j]}$, square relative error $\nicefrac{1}{w \cdot h} \sum_{\boldsymbol{j} \in \Omega_{i}}\nicefrac{(d[j]-\widehat{d}[j])^2}{\widehat{d}[j]}$, root mean square error $(\nicefrac{1}{w \cdot h} \sum_{\boldsymbol{j} \in \Omega_{i}}  (d[j]-\widehat{d}[j])^2 )^{1/2}$, rsme log $(\nicefrac{1}{w \cdot h} \sum_{\boldsymbol{j} \in \Omega_{i}}  (\log d[j]- \log \widehat{d}[j])^2 )^{1/2}$ and $\delta < 1.25^i$: $\nicefrac{1}{w \cdot h} \sum_{\boldsymbol{j} \in \Omega_{i}} \max(\nicefrac{d[j]}{\widehat{d}[j]},\nicefrac{\widehat{d}[j]}{d[j]}) < 1.25^i$. We report also the Area Under the Calibration Error (AUCE) in terms of absolute uncertainty calibration\cite{Gustafsson2019}. Since our methods output a Gaussian distribution $ \mathcal{N}(\widehat{d}, \sigma^2)$ per pixel, we generate prediction intervals ~$\widehat{d} \pm \phi^{-1} (\frac{p+1}{2})\sigma$ of confidence level $p \in [0, 1]$ being $\phi$ the CDF of the standard normal distribution. In a perfectly calibrated model, the proportion of pixels for which the prediction intervals covers the ground truth coincides the confidence level.

\textbf{Synthetic colon dataset. }We evaluate three training alternatives: GT (ground truth) depth supervision, SfM supervision and self-supervision. We use $6,\!550$ images for training and $720$ images for testing. 
We observed that training more than $18$ networks per ensemble does not improve the performance significantly, so we use this number in our experiments.
In SfM-related experiments, we use COLMAP~\cite{schonberger2016structure}. Since $d^{SfM}$ is up to scale, we compute a scale correction factor $s_i$ per image $\mathcal{I}_i$ as follows: $s^{SfM}_i = \nicefrac{\text{median}(d^{GT}_i)}{\text{median}(d^{SfM}_i)}$. This scale correction is also applied to predictions of self-supervised and supervised SfM models that are also up-to-scale.
Table \ref{tab:metricsGT} shows the metrics for the depth error and its uncertainty.
\begin{table}
    \caption{Depth and uncertainty metrics in the synthetic dataset.  RMSE in mm. }
    \resizebox{\columnwidth}{!}{
\begin{tabular}{cccccccccc} Approach  & Abs\textsubscript{Rel} & Sq\textsubscript{Rel}  & RMSE & RMSE\textsubscript{log}  & $\delta < 1.25$  & $\delta < 1.25^2$  & $\delta < 1.25^3$ & AUCE  \\ \hline
 Supervised GT &\textbf{0.050} & \textbf{ 0.335}& \textbf{2.996} &  \textbf{0.102} & \textbf{0.978} & \textbf{0.993} & \textbf{0.997} &  +0.190\\
 
  Supervised SfM & 0.172 & 2.568 & 7.409 & 0.269 & 0.852 & 0.939 & 0.962 & \textbf{-0.116} \\
Self-supervised &  0.179 & 1.774 & 7.601 & 0.243 & 0.792 & 0.938 & 0.972 & -0.152 \\
    \end{tabular}
    }
    \label{tab:metricsGT}
\end{table}

\begin{figure}[ht]
\centerline{\includegraphics[width=\columnwidth]{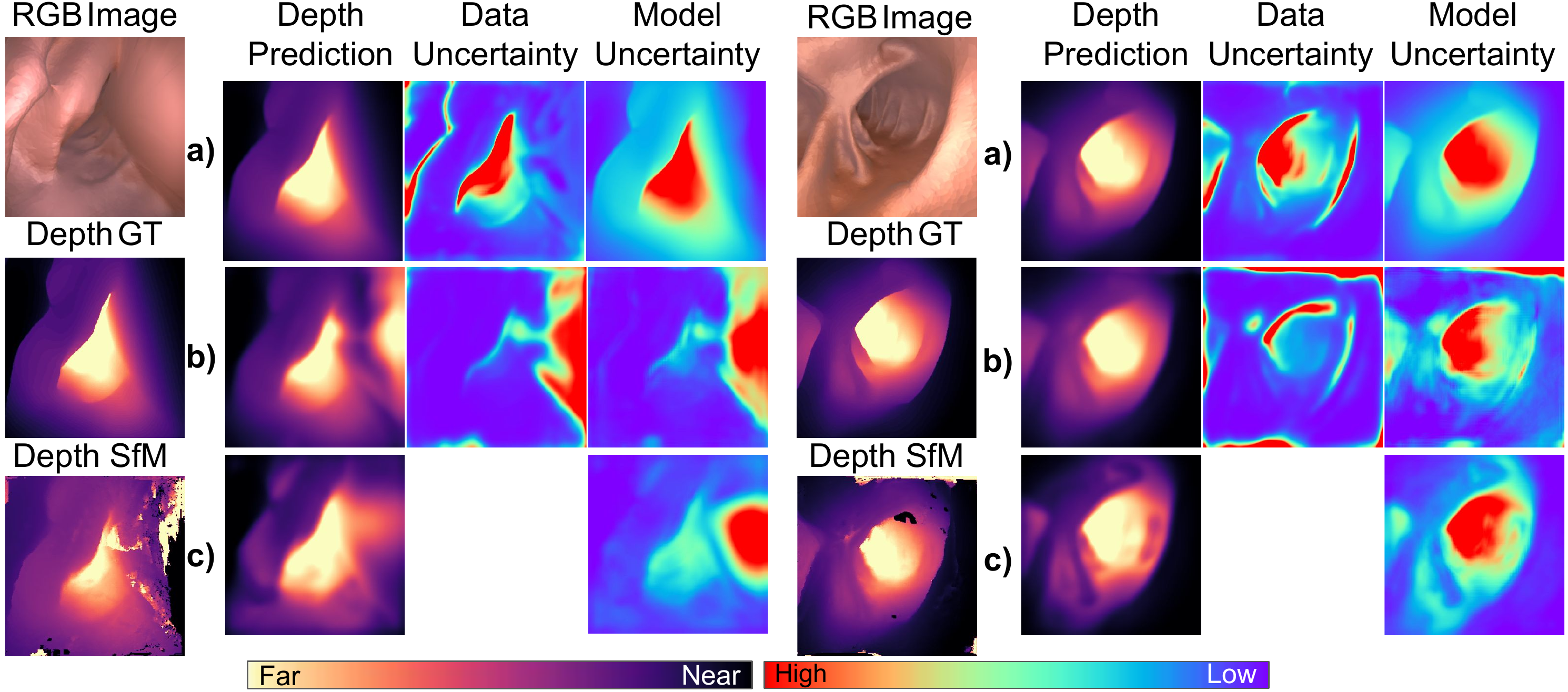}}
\caption{Qualitative depth and uncertainty examples of (supervised learning, supervised learning SfM) and self supervised learning in synthetic images. a) Supervised GT, b) Supervised SfM and c) Self-supervised}
\label{fig:examplesSynth}
\end{figure}

Supervising a deep ensemble with $d^{GT}$ labels achieves the best depth metrics. In terms of uncertainty, self-supervised and supervised with SfM are underconfident, in contrast to supervised with GT depth, which is overconfident and presents higher (worse) absolute AUCE.
Fig.~\ref{fig:examplesSynth} shows that the aleatoric uncertainty supervised by GT is high around the haustras and in dark areas. The epistemic uncertainty grows with the scene depth. The uncertainty supervised by SfM is high in areas where there are typically holes in SfM reconstructions. Similarly to models trained with GT, the aleatoric uncertainty is also visible in the haustras and the epistemic in the deepest areas. 
Photometric self-supervision tends to offer the worst performance.

\textbf{EndoMapper dataset.}
This experiment evaluates Bayesian depth networks in real colonoscopies.
We use the model previously trained with synthetic ground truth depth (``Supervised GT'') to analyse the effect of the domain change. In addition, we also present results from self-supervised training, a baseline teacher-student method \cite{poggi2020uncertainty} and our novel uncertain teacher approach. In real colonoscopies viewpoints change abruptly, images might be saturated or blurry, a considerable amount of liquid might appear and the colon itself produces significant occlusions. For this reasons, we remove images with partial or total visibility issues. We finally use $6,\!912$ images out of the $14,\!400$ images in the complete colonoscopic procedure. In order to obtain depth and uncertainty metrics, we create a 3D reconstruction of the colon using COLMAP (see suplementary material). We also use 18-network ensembles for all methods. Table \ref{tab:real} shows the results. Our ``Uncertain teacher'' shows in general the smallest depth errors and the highest correlation between depth errors and predicted uncertainties.

\begin{table}[t]
    \caption{Depth and uncertainty metrics in the EndoMapper dataset.}
    \centering
    \resizebox{\textwidth}{!}{
\begin{tabular}{ccccccccc} Approach & Abs\textsubscript{Rel}  & Sq\textsubscript{Rel}  & RMSE  & 
RMSE\textsubscript{log}  & $\delta < 1.25$  & $\delta < 1.25^2$  & $\delta < 1.25^3$  & AUCE  \\ \hline
Supervised GT & 0.240 &  0.644 & 2.595 & 0.308 & 0.645 & 0.898 & 0.962 &-0.148\\
Self-supervised & 0.371 & 1.260 & 4.603 & 0.431 & 0.417 & 0.721 &  0.886 & -0.273 \\
Teacher-student & 0.234 & 0.600 & 2.532 & 0.301 & 0.657 & 0.903 & 0.963 &-0.328 \\
Uncertain teacher (ours) & \bf{0.230} & \bf{0.572} & \bf{2.458} & \bf{0.298} & \bf{0.667} & \bf{0.906} &  \bf{0.964} & \bf{-0.129}\\

\end{tabular}
}
    \label{tab:real}
\end{table}

\begin{figure}[t]
\centering
\centerline{\includegraphics[width=\columnwidth]{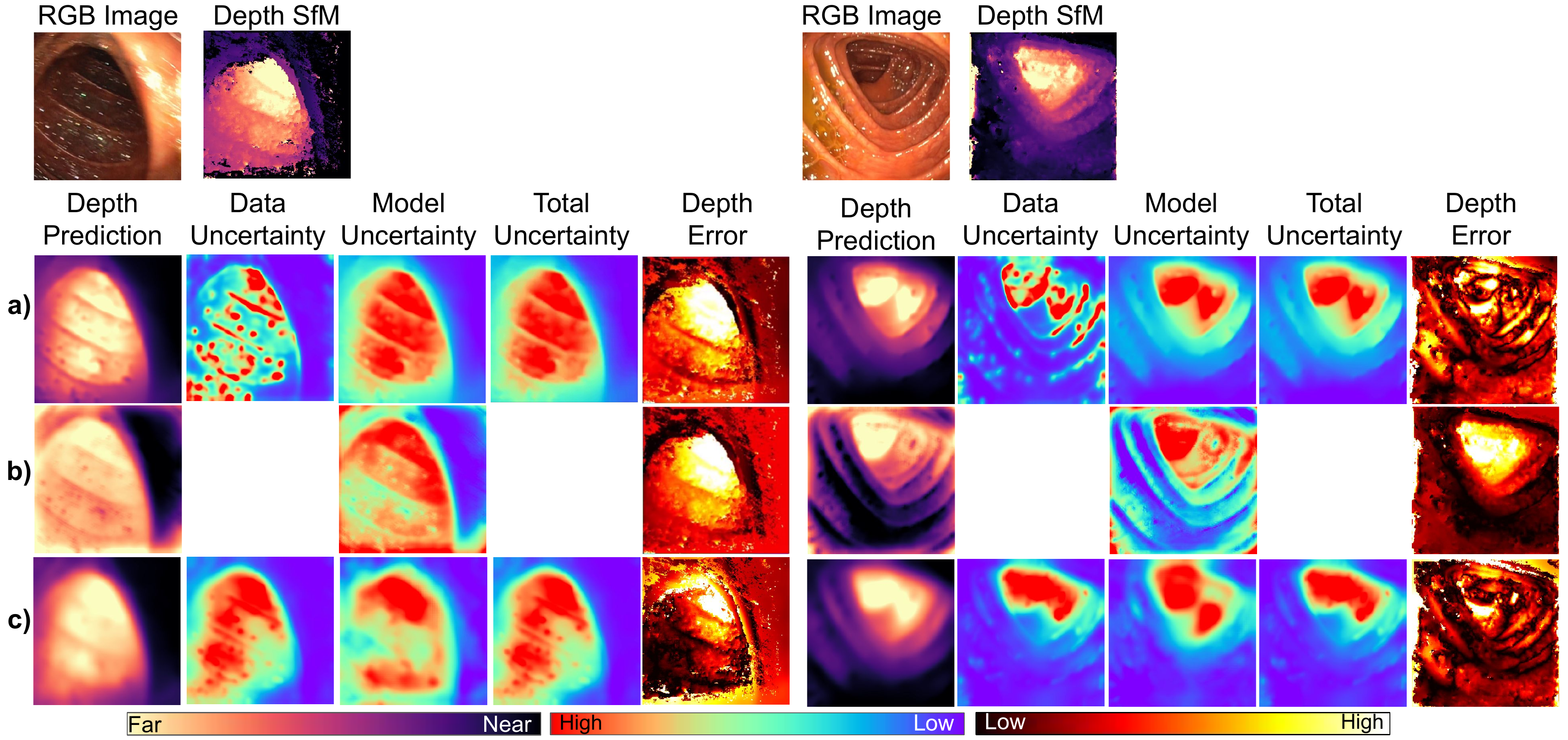}}
\caption{Qualitative depth and uncertainty examples for a) supervised, b) self-supervised and c) uncertain teacher-student learning in real images.
}
\label{fig:examples}
\end{figure}

For self-supervised methods, this real setting is challenging due to reflections, fluids and deformations, all of them aspects that are not considered in the photometric reprojection model of self-supervised losses. ``Supervised GT'' is affected by domain change, as it was trained on synthetic data. However, we observe that it successfully generalizes to the real domain and outperforms the self-supervised method. Based on this observation, we use synthetic supervision in the ``Teacher-student'' baseline and our ``Uncertain teacher''. In general, teacher-student depth metrics outperform the models trained with GT supervision in the synthetic domain and with self-supervision in the real domain. However, ``Teacher-student'' presents the worst AUCE metric, as the teacher uncertainty is not taken into account at training time. Our ``Uncertain teacher'' is the one presenting the best depth and uncertainty metrics, as it appropriately models the noise coming from domain transfer in the depth labels.
Fig. \ref{fig:examples} shows qualitative results for the ``Supervised GT'', ``Teacher-student'' and ``Uncertain teacher'' models. Note that the data uncertainty captures light reflection and depth discontinuities in supervised learning. On the other hand, the model uncertainty grows for the deeper areas. Observing these results, we can conclude that the domain change from synthetic to real colon images is not significant. Models trained on synthetic data generalize to real images and outperform models trained with self-supervision on the target domain, due to the challenges mentioned in the previous paragraphs. 

\section{Conclusions}
\label{sec:Conclusion}
All systems building on depth predictions from color images benefit from uncertainty estimates, in order to obtain robust, explainable and dependable assistance and decisions. In this paper, we have explored for the first time supervised and self-supervised approaches for depth and uncertainty single-view predictions in colonoscopies. 
From our experimental results, we extract several conclusions. Firstly, using ground truth depth as supervisory signal outperforms self-supervised learning and results in better calibrated models. Secondly, approaches based on photometric self-supervision and on SfM supervision coexist in the literature and there is a lack of analysis and results showing which type is more convenient. Thirdly, our experiments show that models trained in synthetic colonoscopies generalize to real colonoscopy images. Finally, we have proposed a novel teacher-student architecture that incorporates the teacher uncertainty in the loss, and have shown that it produces lower depth errors and better calibrated uncertainties than previous teacher-student architectures.
\noindent \subsubsection{Acknowledgments.} This work was supported by EndoMapper GA 863146 (EU-H2020),
RTI2018-096903-B-I00, BES-2016-078426, PID2021-127685NB-I00 (FEDER / Spanish Government), DGA-T45 17R/FSE (Aragón
Government).
\bibliographystyle{splncs04}
\bibliography{biblio}

\end{document}